\documentclass[10pt,twocolumn,letterpaper]{article}

\usepackage{cvpr}
\usepackage{times}
\usepackage{epsfig}
\usepackage{graphicx}
\usepackage{amsmath}
\usepackage{amssymb}

\usepackage[table,xcdraw]{xcolor}
\usepackage{caption} 
\usepackage{algorithm}
\usepackage{algpseudocode}
\usepackage[utf8]{inputenc}
\usepackage{siunitx}

\usepackage{listings}
\usepackage{xcolor}
\usepackage{bm}
 
\definecolor{codegreen}{rgb}{0,0.6,0}
\definecolor{codegray}{rgb}{0.5,0.5,0.5}
\definecolor{codepurple}{rgb}{0.58,0,0.82}
\definecolor{backcolour}{rgb}{0.95,0.95,0.92}
 
\lstdefinestyle{mystyle}{
    backgroundcolor=\color{backcolour},   
    commentstyle=\color{codegreen},
    keywordstyle=\color{magenta},
    numberstyle=\tiny\color{codegray},
    stringstyle=\color{codepurple},
    basicstyle=\ttfamily\footnotesize,
    breakatwhitespace=false,         
    breaklines=true,                 
    captionpos=b,                    
    keepspaces=true,                 
    numbers=left,                    
    numbersep=5pt,                  
    showspaces=false,                
    showstringspaces=false,
    showtabs=false,                  
    tabsize=4
}
 
\algnewcommand{\LineComment}[1]{\State \(\triangleright\) #1}
\usepackage{enumitem}

\newcommand{\dafkld}[2]{{\mathbb{D}({{#1}}\! \mid \! \mid {{#2}})}}
\newcommand{\expect}[1]{{\mathbb{E}}{{\left[{{#1}}\right]}}}
\newcommand{\expee}[2]{{{{{\mathbb{E}}}}_{{#2}}{{\left[{{#1}}\right]}}}}

\newcommand{\var}[1]{{\mbox{var}({#1})}}

\usepackage[pagebackref=true,breaklinks=true,letterpaper=true,colorlinks,bookmarks=false]{hyperref}

\cvprfinalcopy 

\def\cvprPaperID{9530} 

\ifcvprfinal\fi
\begin{document}

\title{Effectively Unbiased FID and Inception Score and where to find them}

\author{Min Jin Chong and David Forsyth \\
University of Illinois at Urbana-Champaign\\
{\tt\small \{mchong6, daf\}@illinois.edu}
}

\maketitle

\begin{abstract}

This paper shows that two commonly used evaluation metrics for generative models, the Fréchet Inception Distance (FID) and the Inception Score (IS), are biased -- the expected value of the score computed for a finite sample set is not the true value of the score. Worse, the paper shows that the bias term depends on the particular model being evaluated, so model A may get a better score than model B simply because model A's bias term is smaller. This effect cannot be fixed by evaluating at a fixed number of samples. This means all comparisons using FID or IS as currently computed are unreliable.  

We then show how to extrapolate the score to obtain an effectively bias-free estimate of scores computed with an infinite number of samples, which we term $\overline{\textrm{FID}}_\infty$ and $\overline{\textrm{IS}}_\infty$. In turn, this effectively bias-free estimate requires good estimates of scores with a finite number of samples.  We show that using Quasi-Monte Carlo integration notably improves estimates of FID and IS for finite sample sets. Our extrapolated scores are simple, drop-in replacements for the finite sample scores. Additionally, we show that using low discrepancy sequence in GAN training offers small improvements in the resulting generator. The code for calculating $\overline{\textrm{FID}}_\infty$ and $\overline{\textrm{IS}}_\infty$ is at \url{https://github.com/mchong6/FID_IS_infinity}.

\end{abstract}

\section{Introduction}
Deep Generative Models have been used successfully to generate hyperrealistic images~\cite{karras2017progressive,karras2019style,brock2018large}, map images between domains in an unsupervised fashion~\cite{zhu2017unpaired,liu2019few}, and generate images from text~\cite{zhang2018stackgan++,xu2018attngan}. Despite their widespread adoption, a simple, consistent, and human-like evaluation of generative models remain elusive, with multiple ad-hoc heuristics having their own faults. The Fréchet Inception Distance (FID)~\cite{heusel2017gans} was shown to have a high bias \cite{binkowski2018demystifying}; Inception Score (IS)~\cite{salimans2016improved} does not account for intra-class diversity and has further been shown to be suboptimal \cite{barratt2018note}; HYPE~\cite{zhou2019hype} requires human evaluation which makes large scale evaluation difficult; Kernel Inception Distance (KID)~\cite{binkowski2018demystifying} has not been widely adopted, likely because of its relatively high variance \cite{ravuri2019seeing}. Write $\textrm{FID}_N$ and IS$_N$ for FID and IS computed with $N$ generated samples. Evaluation procedures with FID vary: some authors use $\textrm{FID}_{50\textrm{k}}$~\cite{brock2018large, karras2019style, karras2017progressive} while others use $\textrm{FID}_{10\textrm{k}}$~\cite{tolstikhin2017wasserstein, lucic2018gans}.

In this paper, we show that both FID and Inception Scores are biased differently depending on the generator. Biased estimators are often preferred over unbiased estimators for the reason of efficiency (a strong application example is the photon map in rendering~\cite{jensen1996global}). In this case however, the bias is intolerable because for both FID and IS, the bias is a function both of $N$ and the generator being tested. This means that we cannot compare generators because each has a different bias term (it is not sufficient to fix $N$, a procedure described in~\cite{binkowski2018demystifying}). To fix this, we propose an extrapolation procedure on $\textrm{FID}_N$ and IS$_N$ to obtain an effectively unbiased estimate $\overline{\textrm{FID}}_\infty$ and $\overline{\textrm{IS}}_\infty$ (the estimate when evaluated with an unlimited number of generated images). In addition, both $\textrm{FID}_\infty$ and $\textrm{IS}_\infty$ are best estimated with low variance estimates of $\textrm{FID}_N$ and $\textrm{IS}_N$. We show that Quasi-Monte Carlo Integration offers useful variance reduction in these estimates. The result is a simple method for unbiased comparisons between models. Conveniently, $\overline{\textrm{FID}}_\infty$ is a drop-in replacement for $\textrm{FID}_N$; $\overline{\textrm{IS}}_\infty$ for Inception Score. Our main contributions are as follows: 

\begin{enumerate}[topsep=1pt,itemsep=0ex,partopsep=1ex,parsep=1ex]
\item We show that $\textrm{FID}_N$ and IS$_N$ are biased, and cannot be used to compare generators.
\item We show that extrapolation of $\textrm{FID}_N$ is reliable, and show how to obtain $\overline{\textrm{FID}}_\infty$ which is an effectively unbiased estimate of FID. Using Quasi-Monte Carlo integration methods yields better estimates of $\overline{\textrm{FID}}_\infty$.
\item We show the same for Inception Score and obtain $\overline{\textrm{IS}}_\infty$, an effectively unbiased estimate of Inception Score.
\item We show that Quasi-Monte Carlo integration methods offer small improvements in GAN training.
\end{enumerate}

All figures are best viewed in color and high resolution.
\section{Background}
\subsection{Fréchet Inception Distance}
To compute Fréchet Inception Distance, we pass generated and true data through an ImageNet~\cite{deng2009imagenet} pretrained Inception V3~\cite{szegedy2016rethinking} model to obtain visually relevant features. Let ($M_t$, $C_t$) and ($M_g$, $C_g$) represent the mean and covariance of the true and generated features respectively, then compute
\begin{equation} \label{eq:fid}
   \textrm{FID} = ||M_t - M_g||_{2}^2 + Tr(C_t + C_g - 2(C_tC_g)^{\frac{1}{2}})
\end{equation}
FID seems to correspond well with human judgement of image quality and diversity \cite{xu2018empirical}.

\subsection{Inception Score}

Write $g(z)$ for an image generator to be evaluated, $y$ for a label, $p(y|x)$ for the posterior probability of a label computed using Inception V3 model on image $x$, $p(y) = \int p(y|g(z)) dz$ for the marginal class distribution, and $\dafkld{p}{q}$ for the KL-divergence between two probability distributions $p$ and $q$.  The Inception Score for a generator is 
\begin{equation}
\exp \left[\expee{\dafkld{p(y|g(z))}{p(y)}}{z \sim p(z)} \right]
\end{equation}

Notice that the marginal class distribution is estimated using the same samples. This is important in our proof that IS is biased. The Inception Score takes into account two properties. 1) Images of meaningful objects should have a conditional label distribution of low entropy. 2) The marginals $p(y)$ should have high entropy if a model is able to generate varied images. A model that satisfy both properties will have a high IS.

\subsection{Monte Carlo and Quasi-Monte Carlo Methods}\label{biasest}
\begin{figure}[t]
    \centering
    \includegraphics[scale = 0.7]{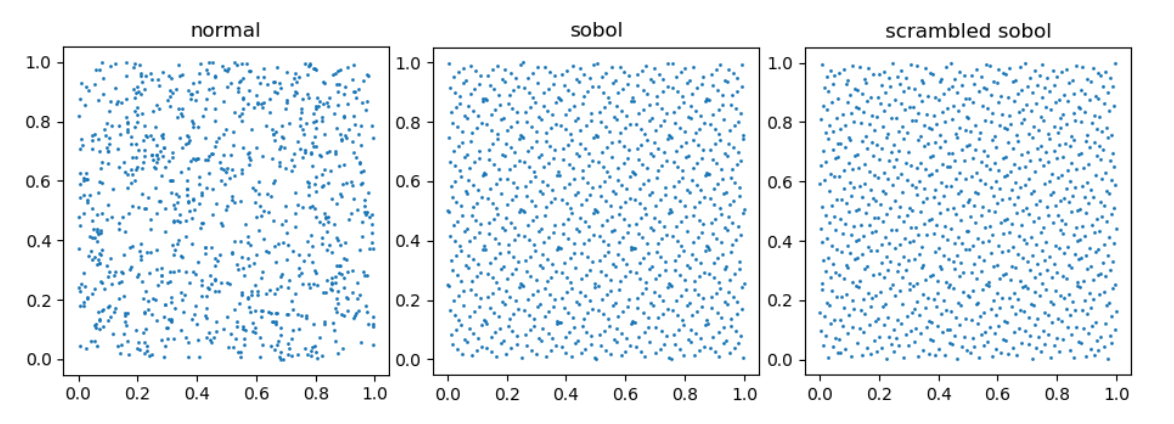}
    \caption{2d scatter plots of $1000$ random points vs Sobol and Scrambled Sobol points. Sobol sequences give us evenly spaced samples while random sampling results in clusters and empty spaces due to their IID property.}
    \label{fig:graph}
\end{figure}

The mean and covariance used in estimating $\textrm{FID}_N$ are Monte Carlo estimates of integrals (the relevant expectations).  The terms $M_t$ and $C_t$ computed on true images are not random, as proper comparisons fix the set of true images used.  However, the terms $M_g$ and $C_g$ are random -- if one uses different samples to evaluate these terms, one gets different values.  A Monte Carlo (MC) estimate of an integral $\int h(x) p(x) dx$ whose true value is $I(h)$, made using $N$ IID samples, yields $\hat{I}=I+\xi$, where 
\begin{equation}
\expect{\xi}=0 \mbox{ and } \var{\xi}=\frac{C(h)}{N}
\end{equation}
where $C(h) \geq 0$ is $\int (h(x)-I(h))^2p(x) dx$~\cite{boyle1977options}.
Note the value of $C$ is usually very hard to estimate directly, but $C$ is non-negative and depends strongly on the function being integrated.
A key algorithmic question is to identify procedures that result in lower variance estimates of the integral. Paskov~\cite{paskov1996faster} showed that Quasi-Monte Carlo Method (QMC) with low-discrepancy sequences such as Sobol~\cite{sobol1967distribution} and Halton~\cite{halton1964algorithm} sequences gave a convergence of up to 5 times faster than MC with lower error rates.  Both MC and QMC approximate
\begin{equation}
    \int_{[0,1]^d}f(u)  du \approx \frac{1}{N}\sum_{i=1}^{N}f(x_i) = \hat{I}
\end{equation}
For MC estimates, $x_i$  are IID samples from uniform distribution on the unit box; for QMC, $x_i$ come from a deterministic quasi-random sequence.  QMC can 
give faster convergence (close to $O(N^{-1})$, compared to MC's $O(N^{-0.5})$~\cite{asmussen2007stochastic}) and lower variance.  This is because IID samples tend to be unevenly spaced (see ``gaps'' and ``clusters'' in Figure~\ref{fig:graph}). QMC points are not independent, and so can be evenly spaced.  A standard QMC construction is the Sobol sequence (review in Dick \etal~\cite{dick2013high}), which forms successively finer uniform partitions of each dimension and then reorders the coordinates to ensure a good distribution. 

\section{Evaluating generative models with FID}

\subsection{FID$_N$ is Biased}\label{FIDbias}

Now consider some function $G$ of a Monte Carlo integral $I(h)$, where $G$ is sufficiently smooth.  We have
\begin{equation}
G(\hat{I})=G(I+\xi)\approx G(I)+\xi G^{'}(I)+\xi^2 \frac{G^{''}(I)}{2}+O(\xi^3)
\end{equation}
so that 
\begin{equation}
\expect{G(\hat{I})}=G(I)+\frac{K}{N}+O(1/N^2)
\end{equation}
where $K=C(h) \frac{G^{''}(I)}{2}$ and $\frac{K}{N}$ is bias.

Consider an estimate of FID, estimated with $N$ samples. The terms $M_g$ and $C_g$ are estimated with an MC integrator, so the estimate must have a bias of $\frac{C_{F}}{N}+O(1/N^2).$  Note that $C_F$ must {\em depend on the generator} $g$ (\autoref{biasest}). Binkowski \etal~\cite{binkowski2018demystifying} note that comparing two generators with different $N$ is unreliable due to bias and that there may be an effect that depends on the generator (but show no evidence). Experiment confirms that (a) $\textrm{FID}_N$ is biased and (b) the bias depends on the generator (Figure~\ref{fig:linear_fid}).

\subsection{$\overline{\textrm{FID}}_\infty$ as an Effectively Unbiased Estimate}\label{sec:unbiased}

The bias in $\textrm{FID}_N$ vanishes for $N \rightarrow \infty$. Figure~\ref{fig:linear_fid} suggests that the $O(1/N^2)$ terms are small for practical $N$, so we can extrapolate in $1/N$ to obtain $\overline{\textrm{FID}}_\infty$ (an estimate of the value of $\textrm{FID}_\infty$). While $\overline{\textrm{FID}}_\infty$ could be still be biased by the higher order terms in the FID$_N$ bias, our experiments suggest these are very small (line fits are good, see Figure~\ref{fig:predict_plot}). Thus, the bias and its dependence on the generator are small and $\overline{\textrm{FID}}_\infty$ is effectively unbiased. While Appendix D.3 of \cite{binkowski2018demystifying} implies there is no estimator of the FID that is unbiased for all distributions for a sample size $N$, our construction removes the very substantial dependence of the bias term on the generator, and so enables comparisons.

However, our extrapolation accuracy depends on the variance of our FID estimates. The estimates are a smooth function $G$ of a Monte Carlo integral $I$. From \autoref{FIDbias}, 
\begin{equation}\label{eq:linear_fid}
    {G(\hat{I})}=G(I)+\frac{K}{N}+O(1/N^2)
\end{equation}
where $K$ depends on $C$ and the first derivative of $G$, and so 
\begin{equation}\label{eq:variance}
\var{G(\hat{I})}=\frac{K_1}{N}+O(1/N^2)
\end{equation}
where $K_1$ depends on $C$ and first and second derivatives of $G$. Notice that this means that an integrator that yields a lower bias estimate of $G(\hat{I})$ will yield a lower variance estimate too for our case (where $G$ is monotonic in $I$). This allows us to identify the integrator to use --- we can find an integrator that yields low variance estimates of $G(I)$ by looking for one that yields the lowest mean value of $G(\hat{I})$. For FID, the best integrator to use is the one that yields the {\em lowest value} of estimated FID, and for IS, the one that yields the {\em highest value} of estimated IS for a given generator. 

Quasi-Monte Carlo methods use low discrepancy sequences to estimate an integral. The Koksma-Hlawka inequality~\cite{owen2003quasi} gives that 
\begin{equation}
\mid I_g-\hat{I}\mid\leq V[f\circ g] D^*_N
\end{equation}
where $V[f\circ g]$ depends on the function to be integrated and is hard to determine, and $D^*_N$ is the discrepancy of the sequence. It is usually difficult to estimate discrepancy, but for Sobol sequences it is $O((log N)^d N^{-1})$ where $d$ is the number of dimensions; for random sequence is $O((\log{\log{N}}/N)^{0.5})$~\cite{owen2003quasi}. As a result, integral estimates with low discrepancy sequences tend to have lower error for the same number of points, though dimension effects can significantly mitigate this improvement. Note that the reduced variance of Sobol sequence estimates manifests as reduced bias (smaller $\textrm{FID}_N$ and larger IS$_N$) and variance in Table~\ref{tbl:fid_tables}. In consequence, Sobol sequence results in better integrators.

{\bf Randomized Sobol sequences:} it is useful to get multiple estimates of the integrals for an FID evaluation because this allows us to estimate the variance of the QMC which helps us construct approximate confidence intervals for the integral. However, low-discrepancy sequences such as the Sobol sequence are deterministic. One way to reintroduce randomness into QMC is to scramble the base digits of the sequence \cite{owen1995randomly}. The resulting sequence will still have a QMC structure 
and the expectation of the integral remains the same.

\subsection{IS$_N$ is Biased}\label{ISbias}

We show the log Inception Score is negatively biased, with a bias term that depends on the generator. Because the exponent is monotonic and analytic, this means the Inception Score is also biased negative with bias depends on the generator. Assuming we have $N$ samples, $X_N = \{x_1, \dots, x_N\}$ with two classes, let $\hat{P}_{1N}=\frac{1}{N}\sum_i p(1|x_i)$ and $p_{1i} = p(1|x_i)$. The log Inception Score over the samples is
\begin{align}
    &\frac{1}{N} \bigg[\sum_i p_{1i}\Big[\log{p_{1i}-\log{\hat{P}_{1N}}}\Big] \\
        +& \sum_i (1-p_{1i})\Big[\log{(1-p_{1i})-\log{(1-\hat{P}_{1N})}}\Big]\bigg]\nonumber\\
        = &\frac{1}{N}\bigg[\sum_i p_{1i}\log{p_{1i}}+ (1-p_{1i})\log{(1-p_{1i})}\bigg]\nonumber\\
        +&\frac{1}{N}\bigg[-\log{\hat{P}_{1N}}\sum_i p_{1i} -\log{(1-\hat{P}_{1N})}\sum_i (1-p_{1i}) \bigg]\nonumber
\end{align}

The first term is a Monte Carlo integral, and so is unbiased. The second term simplifies to the entropy of sample labels. 
\begin{align}
    \frac{1}{N}\bigg[-\hat{P}_{1N}\log{\hat{P}_{1N}} -(1-\hat{P}_{1N})\log{(1-\hat{P}_{1N})} \bigg]
\end{align}

Let $G(u) = -u\log{u} - (1-u)\log{(1-u)}$. By Taylor Series,

\begin{align}
    G(\hat{P}_{1N}) &= G(P_{1N}+\eta) \nonumber\\
    &\approx G(P_{1N}) + \eta\big(\log{(1-P_{1N})}-\log{P_{1N}}\big) \nonumber\\
    &- \frac{\eta^2}{2}\bigg({\frac{1}{P_{1N}(1-P_{1N})}}\bigg) + O(\eta^3)
\end{align}

where $P_{1N}$ is the true integral. When we take expectation over samples, we have
\begin{align}
    G(\hat{P}_{1N}) = G(P_{1N}) - \frac{C}{2N}\bigg(\frac{1}{P_{1N}(1-P_{1N})}\bigg) + O\bigg(\frac{1}{N^2}\bigg)
\end{align}
because $\mathbb{E}[\eta]=0$, $\mathbb{E}[\eta^2]$ as in \autoref{FIDbias}. Note that $C$ must depend on the generator because $p(1|x)$ is shorthand for $p(1|g(z))$. The fact that entropy is convex yields a guaranteed {\em negative} bias in IS as the second derivative of a concave function is non-positive. The multiclass case follows. 

All qualitative features of the analysis for FID are preserved. Particularly, bias depends on the generator (so no comparison with IS$_N$ is meaningful); bias can be corrected by extrapolation (since IS$_N$ is linear w.r.t $\frac{1}{N}$, see Figure~\ref{fig:IS_graph}); and improvements in integrator variance reduce IS$_N$ bias and variance, see Table~\ref{tbl:fid_tables}.

\subsection{Uniform to Standard Normal Distribution}
Low-discrepancy sequences are commonly designed to produce points in the unit hypercube. To make our work a direct drop-in replacement for current generators that use $\mathcal{N}(0, 1)$ as the prior for $z$, we explore two ways to transform a uniform distribution to a standard normal distribution. The main property we want to preserve after the transformation is the low-discrepancy of the generated points.

The Inverse Cumulative Distribution Function (ICDF), gives the value of the random variable such that the probability of it being less than or equal to that value is equal to the given probability. Specifically, 

\begin{equation}
Q(p)={\sqrt {2}}\operatorname {erf} ^{-1}(2p-1),\quad p\in (0,1)
\end{equation}
where $Q(p)$ is the ICDF and erf is the error function. In our case, since our low-discrepancy sequence generates $\mathcal{U}[0,1]$, we can treat them as probabilities and use $Q(p)$ to transform them into $\mathcal{N}(0,1)$.

The Box-Muller transform (BM)~\cite{box1958note}: Given $u \in (0,1)^d$ where $d$ is an even number, let $u^{even}$ be the even-numbered components of $u$ and $u^{odd}$ be the odd-numbered components.
\begin{align}
    \begin{split}
    z_0 &= \sqrt{-2ln(u^{even})}\cos(2\pi u^{odd}) \\
    z_1 &= \sqrt{-2ln(u^{even})}\sin(2\pi u^{odd}) \\
    z &= (z_0, z_1)
    \end{split}
\end{align}

The computations for both methods are negligible, making them efficient for our use case. Okten~\cite{okten2011generating} provided theoretical and empirical evidence that BM has comparable or lower QMC errors compared to ICDF. Our experiments include using both methods, which we dub Sobol$_{BM}$ and Sobol$_{Inv}$. We show that both perform better than random sampling but generally, Sobol$_{Inv}$ gives better estimates for FID$_\infty$ and IS$_\infty$ than Sobol$_{BM}$.

\subsection{Training with Sobol Sequence}

We explore training Generative Adversarial Networks (GANs)~\cite{goodfellow2014generative} with Sobol sequence. GANs are notorious for generating bad images at the tails of the normal distribution where the densities are poorly represented during training. There are several methods such as the truncation trick \cite{brock2018large, karras2019style} which avoids these tail regions to improve image quality at the cost of image diversity. We hypothesize that by using Sobol sequence during GAN training instead of normal sampling, the densities of the distribution will be better represented, leading to overall better generation quality. Furthermore,  we can view training GANs as estimating an integral as it involves sampling a small batch of $z$ and computing the unbiased loss estimate over it. Though we are choosing a small $N$ (batch size in our case), using Quasi-Monte Carlo integration might still lead to a reduction in the variance of the loss estimate.

We note that training a GAN with Sobol sequence has been done before\footnote{\url{https://github.com/deeptechlabs/sobol\_noise\_gan}}. This effort failed because the high-dimensional Sobol points were not correctly generated and were not shuffled. We will describe a successful attempt to train a GAN with Sobol sequence in \autoref{sec:training}.

\section{Experiments}
In our experiments, we find that 
\begin{enumerate}[topsep=1pt,itemsep=0ex,partopsep=1ex,parsep=1ex]
    \item FID is linear with respect to $\frac{1}{N}$ and different generators have very different $K$, so that generators cannot be compared with $\textrm{FID}_N$ for any finite $N$ (\autoref{sec:linear}).
    \item Using Sobol sequence integrators reliably results in lower bias (and so lower variance) in the estimated FID (\autoref{sec:bias}).
    \item Extrapolating the value of $\textrm{FID}_{100k}$ from smaller $N$ compares very well with true estimates. Thus $\textrm{FID}_\infty$ can be estimated effectively with low variance using Sobol points (\autoref{sec:extrapolate} and \autoref{sec:infinity}).
    \item FID$_\infty$ can be estimated effectively for other models such as VAEs~\cite{kingma2013auto} too (\autoref{sec:vae}).
    \item Inception Score behaves like FID but with negative bias. We can estimate $\overline{\textrm{IS}}_N$ accurately (\autoref{sec:inception}).
    \item Training GANs with Sobol sequence yields better $\overline{\textrm{FID}}_\infty$ scores with lower variance across models (\autoref{sec:training}).
\end{enumerate}
Our experiments focus mainly on GANs as they are one of the most popular deep generative models today. We ran our evaluations on DCGAN~\cite{radford2015unsupervised}, ProGAN~\cite{karras2017progressive}, StyleGAN~\cite{karras2019style}, and BigGAN~\cite{brock2018large}. For the implementation of Sobol sequence, we use QMC sampler from BoTorch~\cite{balandat2019botorch}.

We trained a DCGAN on $64 \times 64$ resolution CelebA~\cite{liu2015faceattributes} for $50$ epochs using TTUR~\cite{heusel2017gans} with Adam Optimizer~\cite{kingma2014adam} and Spectral Normalization~\cite{miyato2018spectral}. For ProGAN, we use a pretrained CelebA model for generating $1024 \times 1024$ resolution images. For StyleGAN, we use a pretrained Flickr-Faces-HQ model for generating $1024 \times 1024$ resolution images. We also evaluated on BigGan which is a conditional GAN. We use a pretrained ImageNet BigGAN model which generates $128 \times 128$ resolution images. We also train BigGAN on CIFAR10~\cite{krizhevsky2009learning} and use that for our evaluations.

\subsection{FID$_N$ Bias}\label{sec:linear}
\begin{figure}[t]
    \centering
    \includegraphics[scale = 0.5]{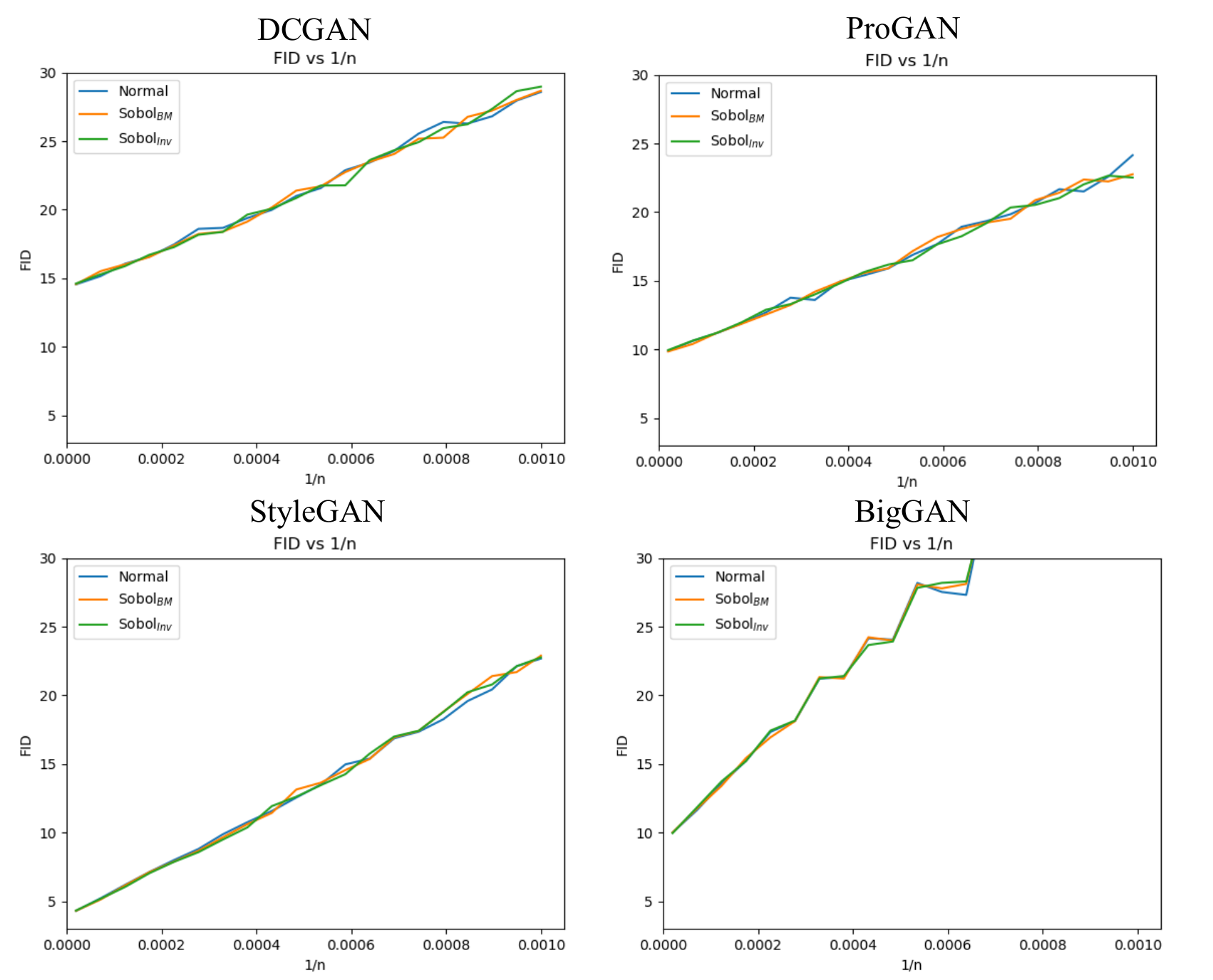}
    \caption{
        Plots of FID vs $\frac{1}{N}$ for various models with various sampling methods at the same scale. Each FID point corresponds to a single FID estimate. FIDs are linear with respect to $\frac{1}{N}$ across all experiments, with higher variance (more spikes) when $N$ is small. Most importantly, the slopes, which corresponds to the $K$ term of eq \ref{eq:linear_fid}, are very different across models. Even though the models are not directly comparable since they generate different datasets, this shows different models have very different $K$ terms. Comparisons of different models with FID$_N$ at fixed $N$ are unreliable because they are dominated by bias.}
    \label{fig:linear_fid}
\end{figure}

\begin{figure}[t]
    \centering
    \includegraphics[scale = 0.6]{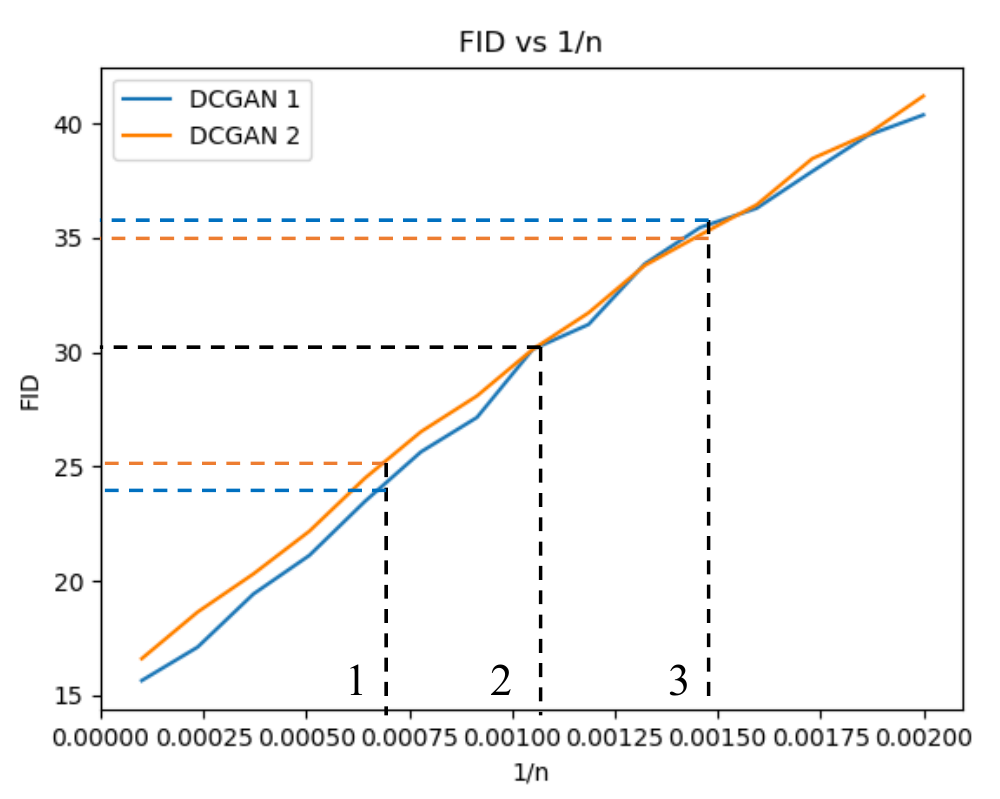}
    \caption{
        The choice of $N$ used affects comparison severely. The graph compares the FID$_N$ vs $\frac{1}{N}$ between two independently trained identical architecture DCGANs. At marker $1$, DCGAN 1 is better than DCGAN 2; at marker 2 they are approximately the same; at marker 3 DCGAN 2 is better than DCGAN 1. This shows that comparisons between models of a fixed $N$ are unreliable.
    }
    \label{fig:compare_fid}
\end{figure}

Across different models, we compare FID at different values of $\frac{1}{N}$ and show that they have a linear relationship as seen in Figure~\ref{fig:linear_fid}. As expected from equation~\ref{eq:variance}, when $N$ is small, the variance of the FID is higher. Importantly, we observe that across different models, the slope varies significantly. The slope corresponds to the $K$ term in equation~\ref{eq:linear_fid} which contributes to the FID bias. In effect, the rankings between GANs are severely dependent on $N$ as different GANs will have different biases that changes with $N$. This can even be seen in two models with the exact same architecture, Figure~\ref{fig:compare_fid}. Appendix D.2 of \cite{binkowski2018demystifying} also gives an empirical example of FID$_N$ reliably giving the wrong ranking of models under somewhat realistic setting. There is no one $N$ that works for every comparison as it depends on the $K$ term of each model. {\em No comparison that uses FID$_N$ is reliable}. 

\subsection{Evaluating with different sampling schemes}\label{sec:bias}
FID is commonly computed with $50$k samples and IS with $5$k samples. Recall \autoref{sec:unbiased} establishes that an integrator that produces a lower bias estimate (which can't be observed) will also produce a lower variance estimate (which can be observed). Table~\ref{tbl:fid_tables} compares $50$ runs each of FID$_{50\textrm{k}}$ and IS$_{5\textrm{k}}$ for a variety of models, estimated using IID normal samples or a Sobol sequence with either Box-Muller transform or ICDF. It is clear from the table that GAN evaluation should always use a low-discrepancy sequence, because evaluating with Sobol$_{BM}$ and Sobol$_{Inv}$ gives a better FID and Inception Score with lower standard deviations.

\begin{table*}[t]
\resizebox{\textwidth}{!}{%
\begin{tabular}{cc|c|c|c|c|c|c|c}
                 & \multicolumn{2}{c}{Normal}                              & \multicolumn{3}{c}{{\color[HTML]{333333} Sobol$_{BM}$}}                                                           & \multicolumn{3}{c}{Sobol$_{Inv}$}                                   \\\hline
Models           & \cellcolor[HTML]{C0C0C0}FID$_{50\textrm{k}}$             & IS$_{5\textrm{k}}$             & \cellcolor[HTML]{C0C0C0} FID$_{50\textrm{k}}$         &  IS$_{5\textrm{k}}$            & F Value & \cellcolor[HTML]{C0C0C0}FID$_{50\textrm{k}}$             & IS$_{5\textrm{k}}$             & F Value \\\hline
DCGAN          & \cellcolor[HTML]{C0C0C0}14.61 $\pm$ 0.0579 & -              & \cellcolor[HTML]{C0C0C0} 14.59 $\pm$ 0.0471 &  -              & 1.51    & \cellcolor[HTML]{C0C0C0}\textbf{14.58 $\pm$ 0.0439} & -              & 1.74    \\
ProGAN           & \cellcolor[HTML]{C0C0C0}9.94 $\pm$ 0.0411  & -              & \cellcolor[HTML]{C0C0C0} \textbf{9.94 $\pm$ 0.0384}  & -              & 1.14    & \cellcolor[HTML]{C0C0C0}9.94 $\pm$ 0.0404  & -              & 1.03    \\
StyleGAN         & \cellcolor[HTML]{C0C0C0}4.33 $\pm$ 0.0413  & -              & \cellcolor[HTML]{C0C0C0} 4.33 $\pm$ 0.0406  &  -              & 1.03    & \cellcolor[HTML]{C0C0C0}\textbf{4.33 $\pm$ 0.0354}  & -              & 1.36    \\\hline
    BigGAN           & \cellcolor[HTML]{C0C0C0}9.94 $\pm$ 0.0564  & 92.96 $\pm$ 2.135 & \cellcolor[HTML]{C0C0C0} \textbf{9.92} $\pm$ 0.0576  & 92.89 $\pm$ 1.961 & 1.19    & \cellcolor[HTML]{C0C0C0}9.93 $\pm$ \textbf{0.0419}  & \textbf{93.21 $\pm$ 1.640} & 1.69    \\
    BigGAN (CIFAR10) & \cellcolor[HTML]{C0C0C0}8.26 $\pm$ 0.0467  & 8.44 $\pm$ 0.1223 & \cellcolor[HTML]{C0C0C0}  \textbf{8.25} $\pm$ 0.0455  & \textbf{8.48} $\pm$ 0.1172 & 1.09    & \cellcolor[HTML]{C0C0C0}8.26 $\pm$  \textbf{0.0446} & 8.45 $\pm$ \textbf{0.1018} & 1.44   
\end{tabular}%
}
\caption{
Using Sobol sequences always give better FID and Inception Score (IS) along with lower standard deviations. The table shows FID$_{50\textrm{k}}$ (lower better) and IS$_{5\textrm{k}}$ (higher better) values of different models evaluated on Normal and Sobol sequences over $50$ runs. The F value is the ratio between the variances of the Normal and Sobol Sequences (the higher it is, the more different their variances are). Bolded values indicate the best score or standard deviation. Better scores implies lower bias which implies lower integrator variance.}
\label{tbl:fid_tables}
\end{table*}

\subsection{FID can be extrapolated}\label{sec:extrapolate}
Using the property that FID is linear with respect to $\frac{1}{N}$, we test the accuracy of estimating FID$_{100\textrm{k}}$ given only $50$k images. We do that by first generating a pool of $50$k images from the generator and randomly sampling them with replacement to compute $15$ FIDs. We then fit a linear regression model over the points, which we can then use for extrapolating $\overline{\textrm{FID}}_{100\textrm{k}}$. Pseudo-code can be found in our Appendix.

We tried two ways of choosing the number of images to evaluate the FID on.
\begin{enumerate}[topsep=1pt,itemsep=0ex,partopsep=1ex,parsep=1ex]
    \item choosing over regular intervals of $N$
    \item choosing over regular intervals of $\frac{1}{N}$
\end{enumerate}
In total, for each of our test models, we run $6$ different experiments, each for $50$ runs. Three ways of sampling $z$ (Normal sampling, Sobol$_{BM}$, Sobol$_{Inv}$) and two ways of choosing $N$ for evaluation. 

\begin{figure}[t]
    \centering
    \hspace*{-.6cm}
    \includegraphics[scale = 0.6]{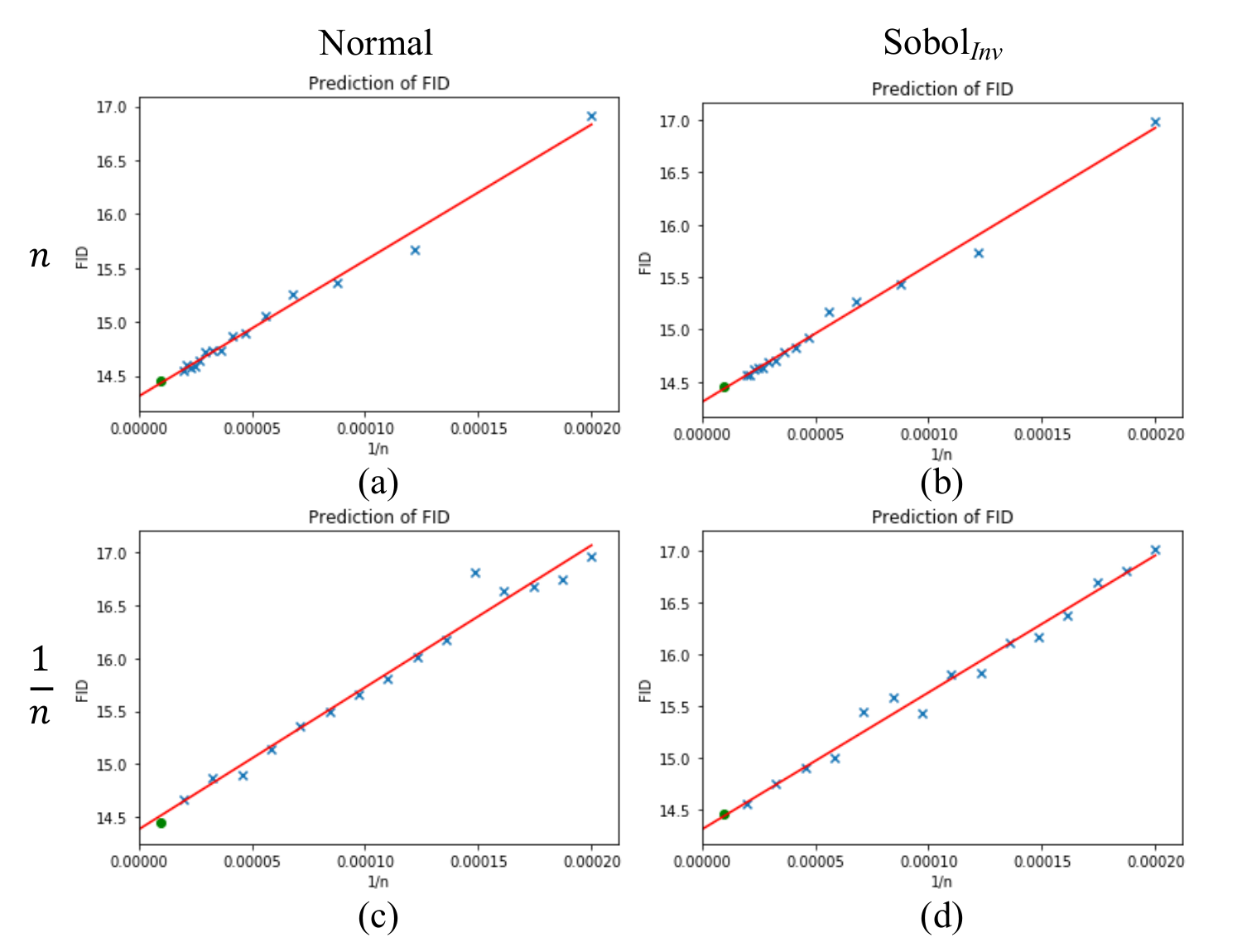}
    \caption{
        FID$_N$ estimates give very good line fits, especially with Sobol$_{Inv}$ (right), suggesting extrapolation will succeed (it does, see Fig \ref{fig:error_plot}). The figure shows the line fits for predicting FID$_{100\textrm{k}}$ for a random DCGAN. Green point is the target FID$_{100\textrm{k}}$ while the blue crosses are the FIDs we compute to fit a linear regression. The columns represents the sampler we use to generate images while the rows represent how we choose $N$ to compute FID. \textbf{Row 1:} choose at regular intervals over $N$; \textbf{Row 2:} choose at regular intervals over $\frac{1}{N}$. For the normal sampler, there are more outliers and the prediction is not as accurate. Using Sobol$_{Inv}$ and computing FIDs at regular intervals over $N$ (Figure \textbf{(b)}) give us better line fit for predicting FID$_{100\textrm{k}}$.}
    \label{fig:predict_plot}
\end{figure}

Computing FID values for $N$ that are evenly spaced in $\frac{1}{N}$ results in an even looking plot, but a weaker extrapolate, see Figure~\ref{fig:predict_plot}. This is because most estimates will be in the region with small $N$, which is noisier. This leads to a poor $\overline{\textrm{FID}}_{100\textrm{k}}$ estimate because the FIDs evaluated at those points have a high variance according to equation~\ref{eq:variance}. Computing the score at regular intervals over $N$ works better in practice. To ensure that the FIDs we calculate are reliable, we use at least $5$k points.

\begin{figure}[t]
    \centering
    \includegraphics[scale = 0.55]{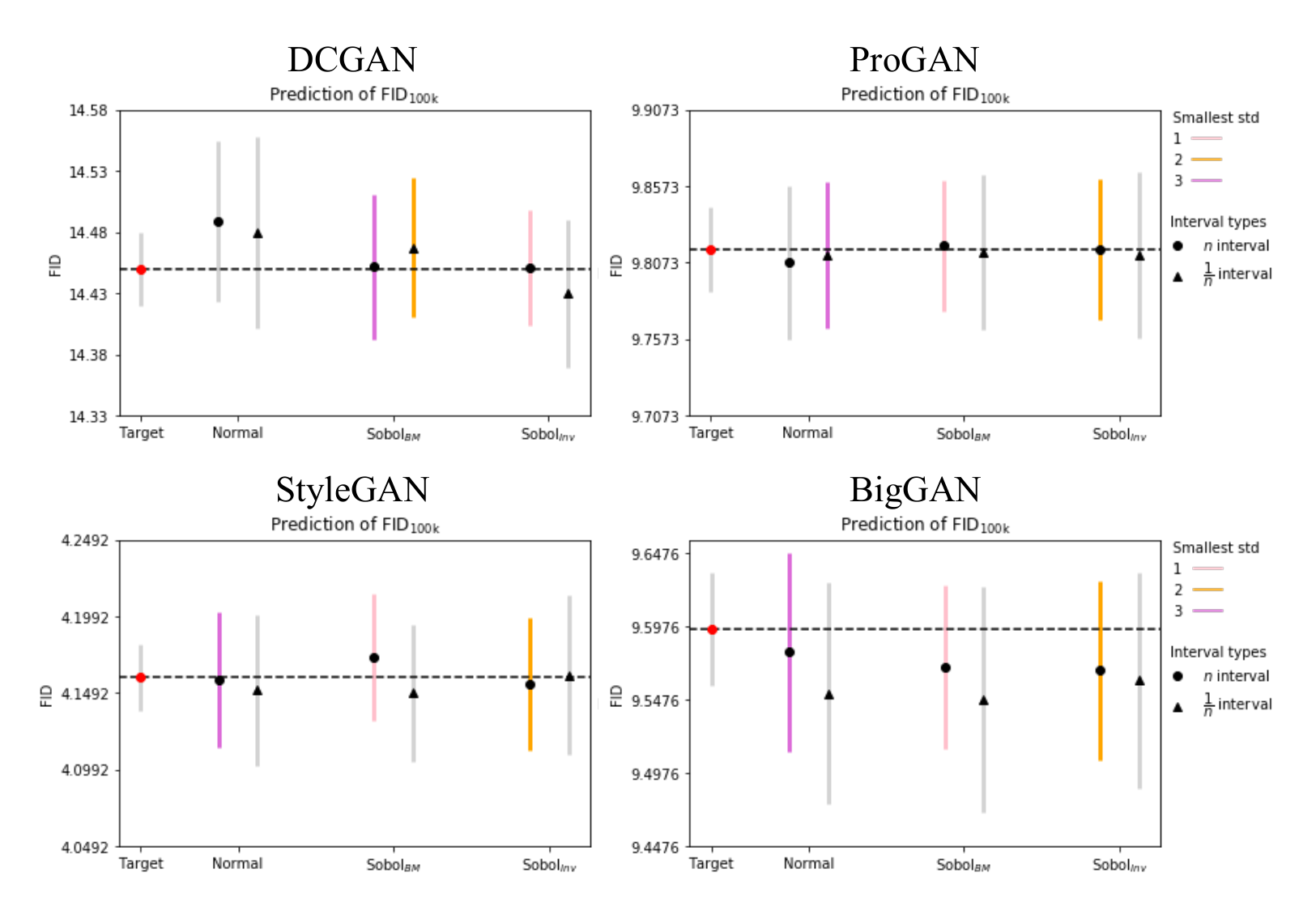}
    \caption{
        FID$_{100\textrm{k}}$ predictions are highly accurate compared to groundtruth (horizontal line) with low variance using $50$k images. This suggests predicting FID$_\infty$ is sound. The figure shows error plots of $\overline{\textrm{FID}}_{100\textrm{k}}$ with $y$ axis of the same scale. The point represent the mean, and the error bar the standard deviation over $50$ runs. The far left point is the target FID$_{100\textrm{k}}$ we are estimating. For each sampling method, we estimate FID$_{100\textrm{k}}$ by fitting points over regular intervals over $N$ (dots) or over $\frac{1}{N}$ intervals (triangles). We also color-tagged the lowest 3 standard deviations. Overall, Sobol$_{Inv}$ with intervals over $N$ perform the best, with good accuracy and low variance. Best viewed in color and high resolution.
    }
    \label{fig:error_plot}
\end{figure}

From Figure~\ref{fig:error_plot}, we can see that across all experiments, $\overline{\textrm{FID}}_{100\textrm{k}}$ is very accurate. Overall, normal random sampling gives a decent estimation but the variance of the estimate is higher compared to using Sobol sequence. Sobol$_{BM}$ has the lowest variance, however, its estimation is not as accurate. Sobol$_{Inv}$ overall gives the best result, giving us an accurate $\overline{\textrm{FID}}_{100\textrm{k}}$ estimate with low variance. This fits into our expectation as FIDs evaluated from Sobol sequence have lower variance, giving us a better line fit, resulting in a more accurate prediction.

More careful tuning of hyperparameters (total number of images and the number of FIDs to fit a line) could yield better $\overline{\textrm{FID}}_{100\textrm{k}}$ estimates.

\subsection{FID$_\infty$}\label{sec:infinity}
\begin{figure}[t!]
    \centering
    \includegraphics[scale = 0.55]{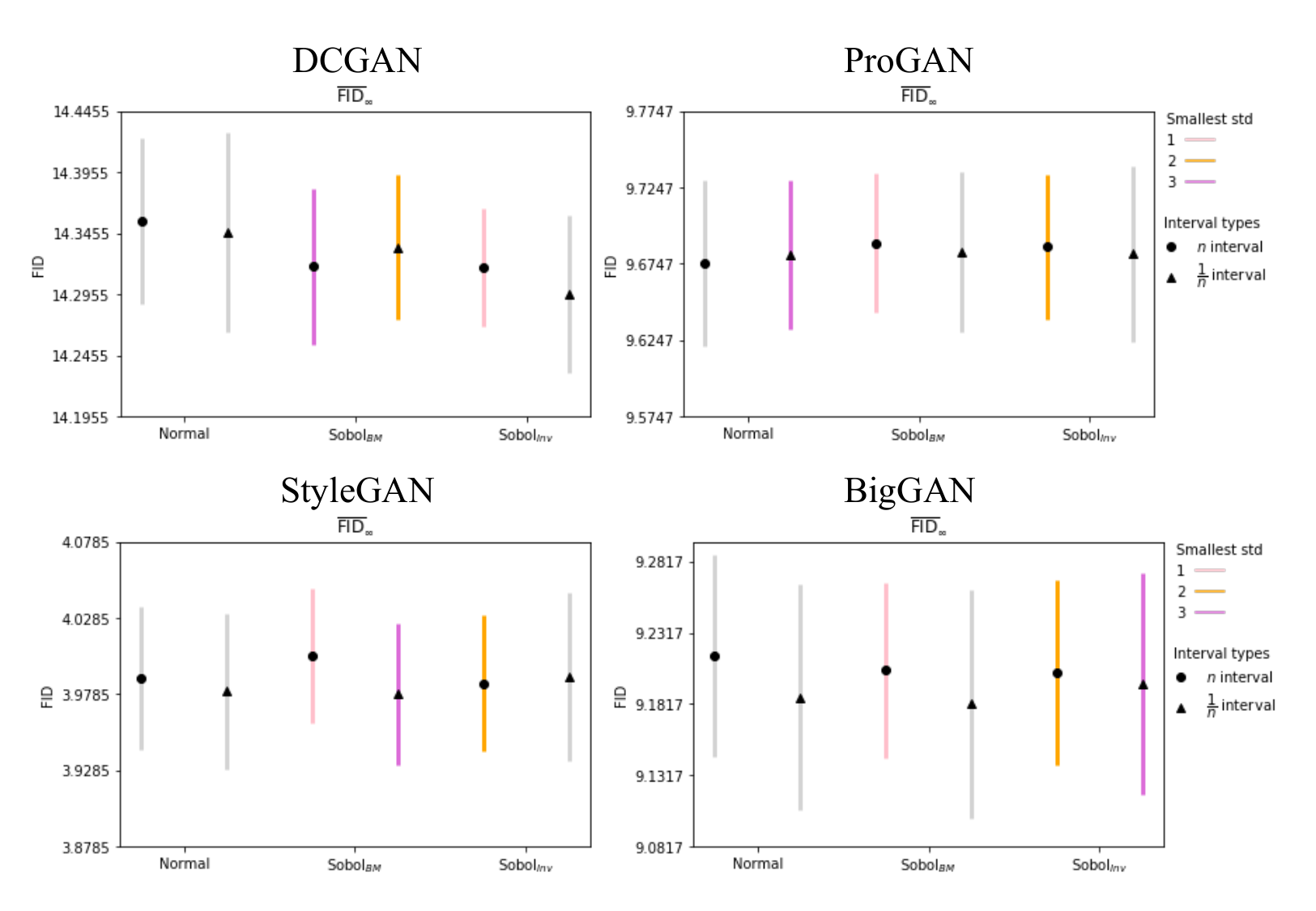}
    \caption{Error plots of predicting FID$_\infty$ over $50$ runs. $\overline{\textrm{FID}}_\infty$ have low variance and are consistent. This together with Figure~\ref{fig:error_plot} suggests that they are accurate. The plots follow the same markings as Figure \ref{fig:error_plot}.
    }
    \label{fig:fid_infinity}
\end{figure}

Since we showed that simple linear regression gives us good prediction accuracy for FID$_{100\textrm{k}}$, we can then extend to estimating FID$_\infty$. Following previous setup, we obtain $\overline{\textrm{FID}}_\infty$ estimate using $50$k samples. Though we do not have the groundtruth FID$_\infty$, our $\overline{\textrm{FID}}_\infty$ estimates (Figure~\ref{fig:fid_infinity}) have similar means across different sampling methods and have small variances. This together with our experiments in \autoref{sec:extrapolate} suggests our $\overline{\textrm{FID}}_\infty$ estimates are accurate.

\subsection{FID$_\infty$ for VAE}\label{sec:vae}
\begin{figure}[t!]
    \centering
    \includegraphics[scale = 0.7]{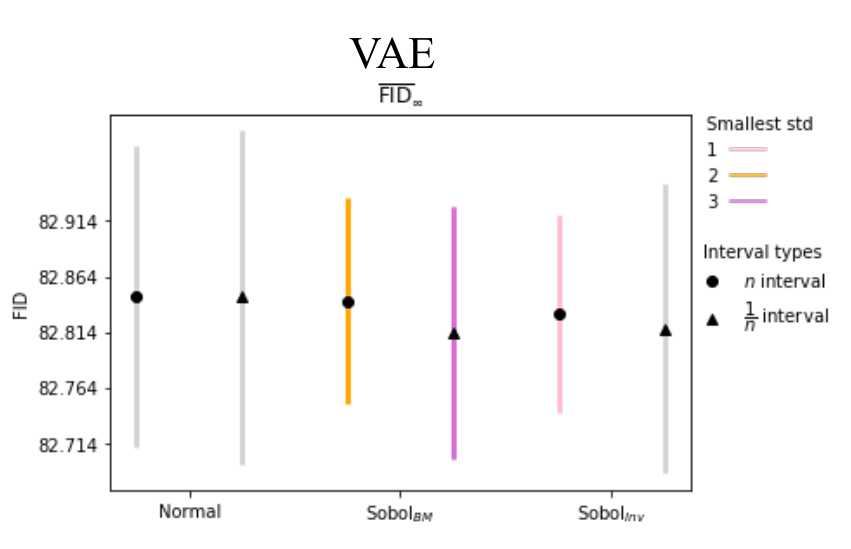}
    \caption{Error plots of predicting FID$_\infty$ for a VAE over $50$ runs. $\overline{\textrm{FID}}_\infty$ works regardless for model used, in this case for VAE. The plots follow the same markings as Figure \ref{fig:error_plot}.
    }
    \label{fig:vae}
\end{figure}

Our FID results apply to any generative model. Our experiments on VAE shows the same linear property of FID, improved bias and variance with QMC, and successful extrapolations. For brevity, we show only the $\overline{\textrm{FID}}_\infty$ plots in Figure~\ref{fig:vae} for a vanilla VAE trained on $64 \times 64$ CelebA.

\subsection{Estimating IS$_\infty$}\label{sec:inception}
\begin{figure}[t!]
    \centering
    \includegraphics[scale = 0.4]{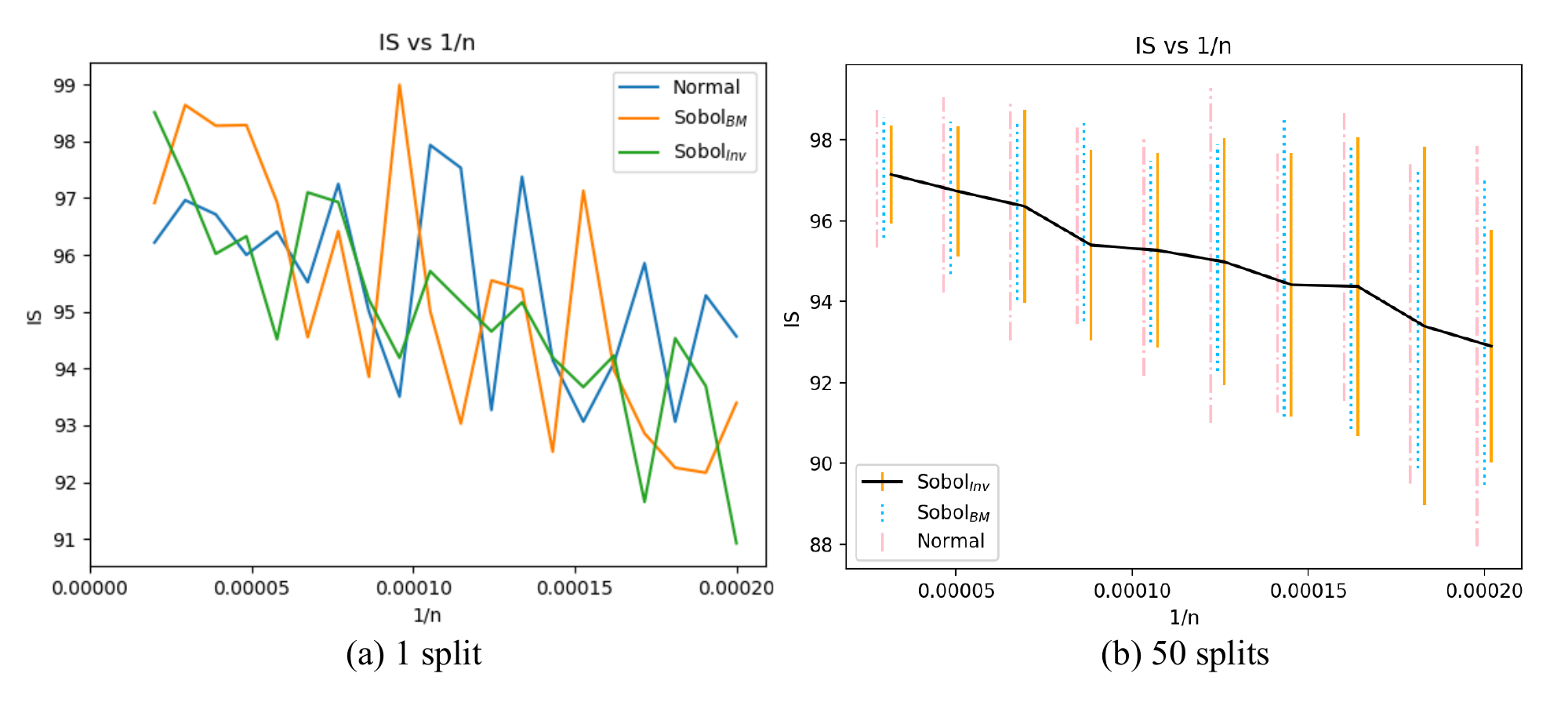}
    \caption{IS$_N$ is negatively biased. IS$_N$ vs $\frac{1}{N}$ of BigGAN for $3$ sampling methods where each point is an average over (a): $1$ split, (b): $50$ splits. For (b), error bars represent standard deviation and the line joins the IS$_N$ estimates mean of Sobol$_{Inv}$. Unlike FID$_N$, IS$_N$ increases with increasing $N$, suggesting negative bias. From (b) the variance of IS$_N$ estimates from normal points are considerably higher than those of Sobol sequences, which is also evidenced by Table~\ref{tbl:fid_tables}. Figure (b) is best viewed in color and high resolution.}
    \label{fig:IS_graph}
\end{figure}

\begin{figure}[t!]
    \centering
    \includegraphics[scale = 0.55]{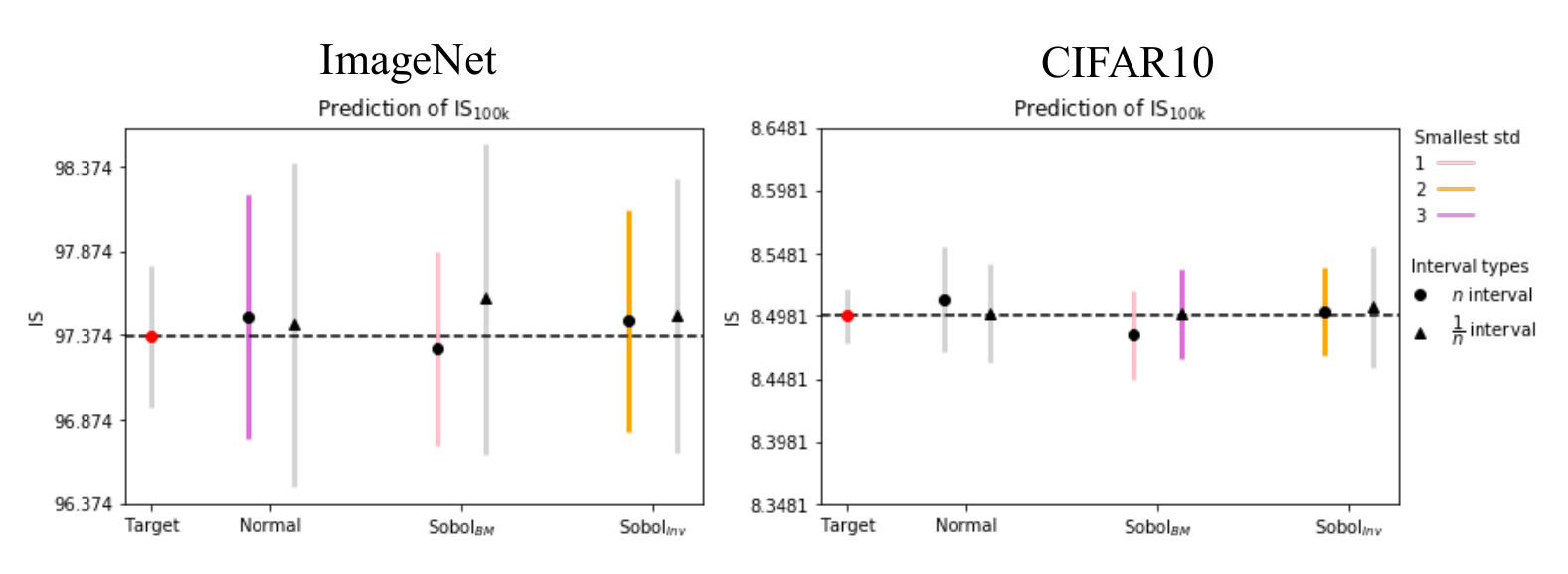}
    \caption{IS$_{100\textrm{k}}$ predictions using $50$k images are highly accurate with low variance. The figure shows error plots of $\overline{\textrm{IS}}_{100\textrm{k}}$ for BigGAN trained on ImageNet and CIFAR10 over $50$ runs. Sobol$_{Inv}$ gives the best accuracy with low standard deviations. The plots follow the same markings as Figure \ref{fig:error_plot}.
    }
    \label{fig:IS_error}
\end{figure}

Inception Score follows the same trend as FID, namely that it is linear with respect to $\frac{1}{N}$ (see Figure~\ref{fig:IS_graph}) and thus can be extrapolated to obtain $\overline{\textrm{IS}}_\infty$ estimate. However, it seems that the variance of IS$_N$ estimates varies greatly for differently generators, see Table~\ref{tbl:fid_tables}. This results in larger variance in our $\overline{\textrm{IS}}_\infty$ estimate which QMC can help reduce, see Figure~\ref{fig:IS_graph}. Figure~\ref{fig:IS_error} shows that the estimated $\overline{\textrm{IS}}_{100\textrm{k}}$ for BigGAN trained on ImageNet is very accurate with comparable variance to the actual IS$_{100\textrm{k}}$ estimate. For CIFAR10 BigGAN, our $\overline{\textrm{IS}}_{100\textrm{k}}$ with Sobol$_{Inv}$ is very accurate with low variance. In general, extrapolating with QMC works and we can get an effectively unbiased estimate $\overline{\textrm{IS}}_\infty$ with good accuracy and low variance.

\subsection{Training with Sobol sequence}\label{sec:training}
We trained DCGAN on CelebA at both $64 \times 64$ and $128 \times 128$ resolution with the same setup as before. For GANs, we need two separate samplers for the generator and discriminator so that the ``uniform'' property of the sequence will not be split between them two. Since Sobol points are highly correlated with each other even with different scramblings, using two Sobol samplers will cause unstable GAN training. Instead, we cache \num{1e6} points for both samplers and shuffle them to break their correlation.

\begin{figure}[t]
    \centering
    \includegraphics[scale = 0.6]{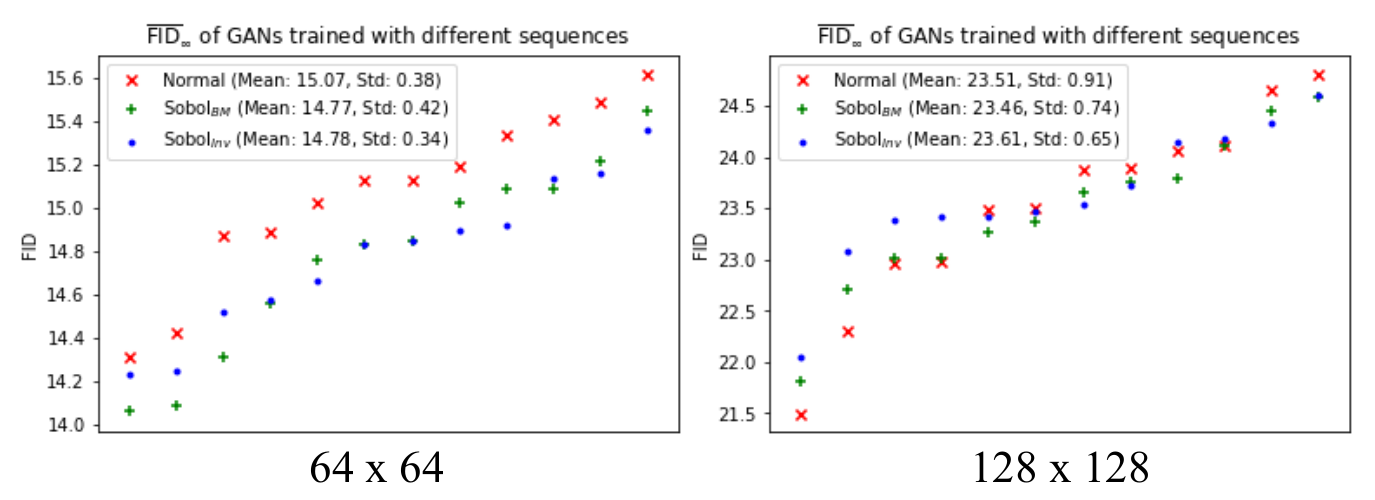}
    \caption{Training GANs with Sobol sequence results in better or comparable $\overline{\textrm{FID}}_\infty$ and lower variance between training runs. The figures show sorted $\overline{\textrm{FID}}_\infty$ of $12$ GANs trained with different sampling methods for two resolutions of CelebA. Each $\overline{\textrm{FID}}_\infty$ is an average of $50$ calculations.}
    \label{fig:train_gan}
\end{figure}

For each of the $3$ sampling methods, we trained $12$ models and we evaluate their $\overline{\textrm{FID}}_\infty$ score over $50$ runs. For $\overline{\textrm{FID}}_\infty$, we use Sobol$_{Inv}$ with regular intervals over $N$. From Figure~\ref{fig:train_gan}, the $\overline{\textrm{FID}}_\infty$ of GANs trained with Sobol sequences are generally lower at $64 \times 64$ and are comparable with normal sampling at $128 \times 128$. However, for both resolutions, GANs trained with Sobol$_{Inv}$ have significantly less $\overline{\textrm{FID}}_\infty$ variance between different runs compared to normal sampling. The improvements are consistent and essentially free as the computational overhead is negligible. We believe further experimentations with more models and datasets could yield interesting results.

\section{Related works}
\textbf{Demystifying MMD GANs:} Binkowski \etal~\cite{binkowski2018demystifying} showed that there is no unbiased estimator for FID. However, the Stone–Weierstrass theorem allows arbitrarily good uniform approximation of functions on the unit interval by sufficiently high degree polynomials. Thus, while zero bias is unattainable, very small bias is not ruled out. Our $\overline{\textrm{FID}}_\infty$ and $\overline{\textrm{IS}}_\infty$ scores clearly display very small bias because (as the graph shows) the $\frac{1}{N}$ terms dominate higher order terms.

\textbf{Debiasing via Importance Weighting:} Grover \etal~\cite{grover2019bias} reduce errors in MC estimates computed with augmented datasets by using a classifier to estimate importance weights. By doing so, they exhibit improved IS$_N$, FID$_N$, and KID$_N$ scores. We believe the improvements are the result of increased effective sample size from the augmentation. However, in contrast to our work, they do not identify formal statistical bias in FID$_N$ or IS$_N$, nor do they point out that the dependence of this bias on the generator makes comparisons at fixed $N$ unreliable.

\textbf{QMC Variational Inference:} Buchholz \etal~\cite{buchholz2018quasi} propose using QMC to reduce the variance of the gradient estimators of Monte Carlo Variational Inference. In their Appendix D, they suggest using QMC for VAEs and GANs. However, they offer no explanations nor results from doing so. 

\textbf{HYPE:} Note that \cite{zhou2019hype} computes a correlation between HYPE and FID$_N$ for generators trained on different ImageNet classes. Because the FID$_N$ scores are biased, {\em with a bias that depends on the particular generator}, the correlations cannot be relied upon. It would be interesting to correlate FID$_\infty$ with HYPE.

\section{Future work}
This paper serves as an introduction of using Quasi-Monte Carlo methods to estimate high dimensional integrals in the field of generative models for bias reduction. However, there has been a substantial amount of work in the area of estimating high dimensional integrals such as sparse grids~\cite{smolyak1963quadrature}, higher order scrambled digital nets~\cite{dick2011higher}, randomized lattice rules~\cite{joe1990randomization} which we have yet to touch upon. Furthermore, using a closed quasi-random sequence (where we know $N$ beforehand) for evaluation could give us better error bounds on the integral \cite{dick2013high}. We reserve these for our future work. Also, FID$_\infty$ could well correlate with HYPE and we plan to investigate these correlations.

\section{Best practices}
\subsection{For evaluating generators with FID or IS}
Never compare generators with FID$_N$ or IS$_N$; the comparisons are not reliable.
\begin{enumerate}[topsep=1pt,itemsep=0ex,partopsep=1ex,parsep=1ex]
    \item Use Sobol$_{Inv}$ to compute both FID$_N$ and IS$_N$.
    \item Using estimates obtained at regular intervals over $N$, extrapolate to get $\overline{\textrm{FID}}_\infty$ and $\overline{\textrm{IS}}_\infty$ estimates.
    \item Repeat multiple times to get a variance estimate. This corresponds to how reliable the $\overline{\textrm{FID}}_\infty$ and $\overline{\textrm{IS}}_\infty$ estimates are.
\end{enumerate}
\subsection{For training GANs}
We have moderate results to show that training with Sobol sequence results in better or comparable $\overline{\textrm{FID}}_\infty$ and lower variance across models. We believe that large-scale experiments should be done to validate the usefulness of training with Sobol sequence and that is left to future work.

{\small
\bibliographystyle{ieee_fullname}
\bibliography{egbib}
}
\clearpage
\newpage
\mbox{}
\section*{Appendix}
Pseudo-code for calculating $\overline{\textrm{FID}}_\infty$. Evaluating $\overline{\textrm{IS}}_\infty$ is exactly the same except swapping the FID calculation function for IS calculation function.
\begin{algorithm}
\caption{Evaluating $\overline{\textrm{FID}}_\infty$ for a Generator }
\begin{algorithmic}[1]
\State $G$: Generator
\State $I$: Inception Network
\State $n$: Number of images we want to generate
\State $t$: Number of FIDs we compute for extrapolation
\State $m_t$: Precomputed groundtruth activations mean
\State $c_t$: Precomputed groundtruth activations covariance 
\State 
\Procedure{Evaluate}{$G, I, n, t, m_t, c_t$}
    \State FID $= \{\}$
    \State $z \gets \textsc{SobolSampler($n$)}$
    \State activations $\gets I(G(z))$
    \State
    \LineComment $5000$ is the min no. of images we evaluate with
    \State batchSizes = $\textsc{linspace($5000, n, t)$}$
    \For{batchSize in batchSizes}
        \State \textsc{Shuffle(\textnormal{activations})}
        \State activations$_i$ $\gets$ activations[:batchSize]
        \State FID.insert(\textsc{CalculateFID(\textnormal{activations$_i$, $m_t, c_t$})})
    \EndFor
    \State reg $\gets \textsc{LinearRegression(1/\textnormal{batchSizes, FID})}$
    \State $\overline{\textrm{FID}}_\infty \gets$ reg.predict($0$)
    \State \textbf{return} $\overline{\textrm{FID}}_\infty$
\EndProcedure
\end{algorithmic}
\end{algorithm}

\begin{lstlisting}[language=Python, caption=PyTorch code for generating Sobol points, float=*]
from botorch.sampling.qmc import NormalQMCEngine

class randn_sampler():
    """
    Generates z~N(0,1) using random sampling or scrambled Sobol sequences.
    Args:
        ndim: (int)
            The dimension of z.
        use_sobol: (bool)
            If True, sample z from scrambled Sobol sequence. Else, sample 
            from standard normal distribution.
            Default: False
        use_inv: (bool)
            If True, use inverse CDF to transform z from U[0,1] to N(0,1).
            Else, use Box-Muller transformation.
            Default: True
        cache: (bool)
            If True, we cache some amount of Sobol points and reorder them.
            This is mainly used for training GANs when we use two separate
            Sobol generators which helps stabilize the training.
            Default: False
            
    Examples::

        >>> sampler = randn_sampler(128, True)
        >>> z = sampler.draw(10) # Generates [10, 128] vector
    """

    def __init__(self, ndim, use_sobol=False, use_inv=True, cache=False):
        self.ndim = ndim
        self.cache = cache
        if use_sobol:
            self.sampler = NormalQMCEngine(d=ndim, inv_transform=use_inv)
            self.cached_points = torch.tensor([])
        else:
            self.sampler = None

    def draw(self, batch_size):
        if self.sampler is None:
            return torch.randn([batch_size, self.ndim])
        else:
            if self.cache:
                if len(self.cached_points) < batch_size:
                    # sample from sampler and reorder the points
                    self.cached_points = self.sampler.draw(int(1e6))[torch.randperm(int(1e6))]

                # Sample without replacement from cached points
                samples = self.cached_points[:batch_size]
                self.cached_points = self.cached_points[batch_size:]
                return samples
            else:
                return self.sampler.draw(batch_size)
\end{lstlisting}

\end{document}